\documentclass[a4paper,twoside]{article}

\usepackage{epsfig}
\usepackage{subcaption}
\usepackage{calc}
\usepackage{amssymb}
\usepackage{amstext}
\usepackage{amsmath}
\usepackage{amsthm}
\usepackage{multicol}
\usepackage{pslatex}
\usepackage{natbib}
\usepackage{algorithm2e}
\usepackage[bottom]{footmisc}
\usepackage{graphicx}
\usepackage{booktabs}
\usepackage{subcaption}

\usepackage{float}
\usepackage{xurl}
\usepackage{hyperref}
\usepackage{SCITEPRESS}

\usepackage{tcolorbox}
\tcbuselibrary{skins,breakable}

\newtcolorbox{prompttemplate}[1]{
  enhanced, colback=white, colframe=black!60,
  arc=1.5mm, boxrule=0.4pt, top=4mm, left=2mm, right=2mm, bottom=2mm,
  fonttitle=\bfseries\small, fontupper=\ttfamily\scriptsize,
  attach boxed title to top center={yshift=-2.5mm},
  boxed title style={colback=#1, colframe=#1, boxrule=0pt, arc=1mm,
                      left=3mm, right=3mm, top=0.8mm, bottom=0.8mm},
  title
}

\newcommand{\promptheader}[1]{\textit{\textbf{\# #1}}\\[1mm]}

\begin{document}

\title{Sizing Up Vision-Language Models: Metric Size Estimation and Comparison on Real-World Product Images}

\author{\authorname{Richardo Serban\sup{1}\orcidAuthor{0009-0005-5284-842X}, Moritz Alker\sup{1}\orcidAuthor{0009-0005-5982-9459}, Andreas Stöckl\sup{1}\orcidAuthor{0000-0003-1646-0514}, Rüdiger Brecht\sup{2}\orcidAuthor{0000-0003-4040-3122}, Michael Weberberger\sup{2} and David C. Schedl\sup{1}\orcidAuthor{0000-0002-7621-3526}}
\affiliation{\sup{1}Digital Media Lab, University of Applied Sciences Upper Austria, Hagenberg, Austria}
\affiliation{\sup{2}Impact Lab, Premedia GmbH, Wels, Austria}
\email{\{andreas.stoeckl, david.schedl\}@fh-hagenberg.at, richardo.serban@gmail.com, moritz.alker@outlook.de, ruediger.brecht@premedia.at, michael.weberberger@mundm-consulting.at}
}


\abstract{%
Estimating real-world object dimensions from a single image is a long-standing computer vision challenge with practical relevance to e-commerce, interior design, and product visualization. Vision-language models (VLMs) are increasingly used for such perception tasks, yet whether they can recover metric size accurately, rather than merely plausibly, remains open. We investigate this using a curated, manually verified 4{,}725-item subset of the Amazon Berkeley Objects (ABO) dataset, benchmarking four VLMs on direct three-axis size estimation with native reasoning disabled where the provider's API permits it. All four models clearly outperform a category-median baseline that knows only the product type (8--19 cm vs.\ 26 cm mean absolute error on the longest side), so their estimates are not merely category priors. Comparing a product-cutout-only condition against one adding scene context, we find scene context consistently improves accuracy, reducing MAE and MARE in every case, with the largest gains for product types whose scale is ambiguous from the cutout alone. A separate sweep of elaborate prompting strategies yielded no benefit beyond run-to-run noise. Scene context, not prompt engineering, is the more reliable lever for grounding VLM size estimates in visual evidence.}

\onecolumn \maketitle \normalsize \setcounter{footnote}{0} \vfill

\section{\uppercase{Introduction}}
\label{sec:introduction}
Knowing the real-world size of an object from a single image is a fundamental visual skill that humans exercise almost effortlessly, yet it remains difficult to automate. Augmented reality systems must place virtual content at a plausible scale, robots must judge whether an object fits through a gap, and automated content pipelines must decide whether a generated or composited object looks physically correct next to its surroundings. Vision-language models (VLMs) are increasingly used as general-purpose perception modules in such pipelines, prized for their semantic understanding and instruction following. Their capacity for metric, quantitative perception, however, is far less well established: a model may correctly recognize “a dining chair“ while being off by tens of centimeters on its height, an error that is invisible to captioning or VQA metrics but critical for any use that depends on physical plausibility.

Two questions motivate this work. First, how accurately can current, off-the-shelf VLMs estimate the real-world size of everyday objects from a single product image, without task-specific fine-tuning or auxiliary depth input? Second, does additional visual context -- a milieu image showing the object in a real scene alongside the plain product cutout -- measurably improve the estimates, or do they depend mainly on category-level priors regardless of context? The second question has practical weight: milieu imagery is not always available at inference time, so if scene context provides no reliable benefit, a cutout-only pipeline is preferable on cost and simplicity grounds.

We study these questions on a benchmark derived from the Amazon Berkeley Objects (ABO) dataset~\citep{collins2022abo}, which provides catalog product images together with physical dimensions. Models estimate the object's width, length, and height in centimeters; we evaluate the predicted \emph{longest side} (and the volume) against the catalog extents, a reduction that is invariant to axis-naming conventions. Our contributions are:
\begin{itemize}
  \item a curated, relabeled evaluation set of 4{,}725 ABO products with verified size annotations, released as a replication kit (download script and index);
  \item a controlled comparison of direct 3D size estimation with and without milieu scene context across four commercial VLMs, with native reasoning disabled wherever the provider's API permits it;
  \item an analysis of the error characteristics across models and object categories, including an image-free category-median baseline and a per-category breakdown of the scene-context effect.
\end{itemize}

\section{\uppercase{Related Work}}
\label{sec:relatedwork}
Despite their semantic competence, VLMs perform close to chance on elementary spatial relations \citep{kamath2023whatsup,liu2023vsr}, struggle with core perceptual tasks such as relative depth ordering \citep{fu2024blink}, and fail systematically on synthetic depth and height probes \citep{azad2025geometer}. Pre-LLM work already targeted size commonsense explicitly \citep{bagherinezhad2016elephants,liu2022spatialcommonsense}. A second line of work \emph{improves} spatial capabilities with 3D-grounded training data and depth input: SpatialVLM \citep{chen2024spatialvlm}, SpatialRGPT \citep{cheng2024spatialrgpt}, SpatialBot \citep{cai2025spatialbot}, MM-Spatial \citep{daxberger2025mmspatial}, SD-VLM with the 700K-pair MSMU dataset \citep{chen2025sdvlmspatialmeasuringunderstanding}, and VLM-3R, which adds a metric-scale geometry encoder for monocular video \citep{fan2026vlm3rvisionlanguagemodelsaugmented}. VSI-Bench \citep{11094671} finds spatial reasoning, rather than perception or language, to be the main bottleneck of current MLLMs, and that chain-of-thought or self-consistency do not improve spatial answers; SPHERE \citep{zhang-etal-2025-sphere} reaches a similar conclusion for single images. Closest to our setting, \citet{meivar2026illposeddesignprobingevidence} use monocular metric size estimation as a deliberately ill-posed diagnostic and find that even the largest open-weight VLMs rely primarily on the identity of the target object and make little use of scene geometry. \citet{liao2024qspatial} show on Q-Spatial Bench that prompting models to reason via visible \emph{reference objects} improves metric estimates without fine-tuning, and \citet{alker2025hail} measure hailstones from crowdsourced images with a two-stage reference-object prompt. Classical single-image metrology instead relies on scene priors \citep{hoiem2008putting}, learned metric depth \citep{bhat2023zoedepth,yang2024depthanythingv2}, or monocular 3D detection \citep{brazil2023omni3d}, all of which need scene context or calibration cues that product photography on a white background does not offer. Unlike these approaches, we evaluate frozen, commercial VLMs on single static product images, closer to the constraints of real product photography.

Prompting is the primary lever for eliciting reasoning from frozen models: chain-of-thought \citep{wei2022cot}, self-consistency \citep{wang2023selfconsistency}, Multimodal-CoT \citep{zhang2024multimodalcot}, Set-of-Mark \citep{yang2023som}, and few-shot in-context learning \citep{alayrac2022flamingo}; see \citet{sahoo2024prompt} and \citet{wu2024visualprompting} for surveys. Most of these studies evaluate prompting either on models without a distinct reasoning mode or with native reasoning left at its default, so the interaction between prompt structure and a model's internal reasoning is rarely isolated. We address this gap by disabling native reasoning wherever the API permits it, and we discuss in Section~\ref{sec:prelim-exploration} why we did not adopt elaborate reference-object prompt variants. On the data side, ShapeNet \citep{chang2015shapenet} and Objaverse \citep{deitke2023objaverse} provide 3D assets largely without metric scale, Pix3D \citep{sun2018pix3d} aligns real furniture images with 3D shapes, and ABO \citep{collins2022abo} is unique in combining $\sim$148k real product listings, with catalog and in-context images, with physical dimensions, which makes it a natural testbed for the setting we adopt.
\section{\uppercase{Methodology}}
\label{sec:method}
We formulate size estimation as a three-dimensional regression task: given a target image of a product on a white background, optionally accompanied by a milieu image showing that product in a real scene, a model estimates the real-world width, length, and height of the object in centimeters. The sole experimental factor in our main sweep is the presence or absence of the milieu image (Section~\ref{sec:prompt-construction}); an earlier, more elaborate sweep over prompt components is reported as a negative result in Section~\ref{sec:prelim-exploration}.

\subsection{Dataset Adjustments and Relabeling}
\label{sec:dataset-adjustments}
ABO~\citep{collins2022abo} provides catalog metadata including physical dimensions, but the raw annotations are not directly usable as ground truth: the three extents are frequently mislabeled with respect to the true width/length/height axes of the assembled product. We sidestep this at evaluation time with two permutation-invariant reductions of the three dimensions, the \emph{longest side} (maximum of the three predicted values vs.\ maximum of the three annotated extents) and the \emph{volume} (product vs.\ product). Both are invariant to axis-labeling conventions by construction, while still requiring the model to reason about the full 3D extent. In addition, we applied the following filtering and relabeling.

\paragraph{Filtering and Manual Verification.}
Dimension entries in ABO stem from heterogeneous catalog fields with mixed units and free-text formats. We retained only listings with correctly specified, structured dimensions and removed listings whose values were on the wrong axis or only in free text, multi-item bundles whose dimensions refer to the packaging, products with variant-dependent or adjustable sizes (e.g.\ height-adjustable chairs, extendable tables), and entries with physically implausible values. The remaining pool of several thousand candidates was then reviewed manually in three passes, verifying for each listing that the target and milieu image depict the same physical product and that the listed dimensions, converted to centimeters where necessary, are consistent with the depicted object.

\begin{figure}[!ht]
  \centering
  \begin{subfigure}[b]{0.49\linewidth}
    \centering
    \includegraphics[width=3.6cm]{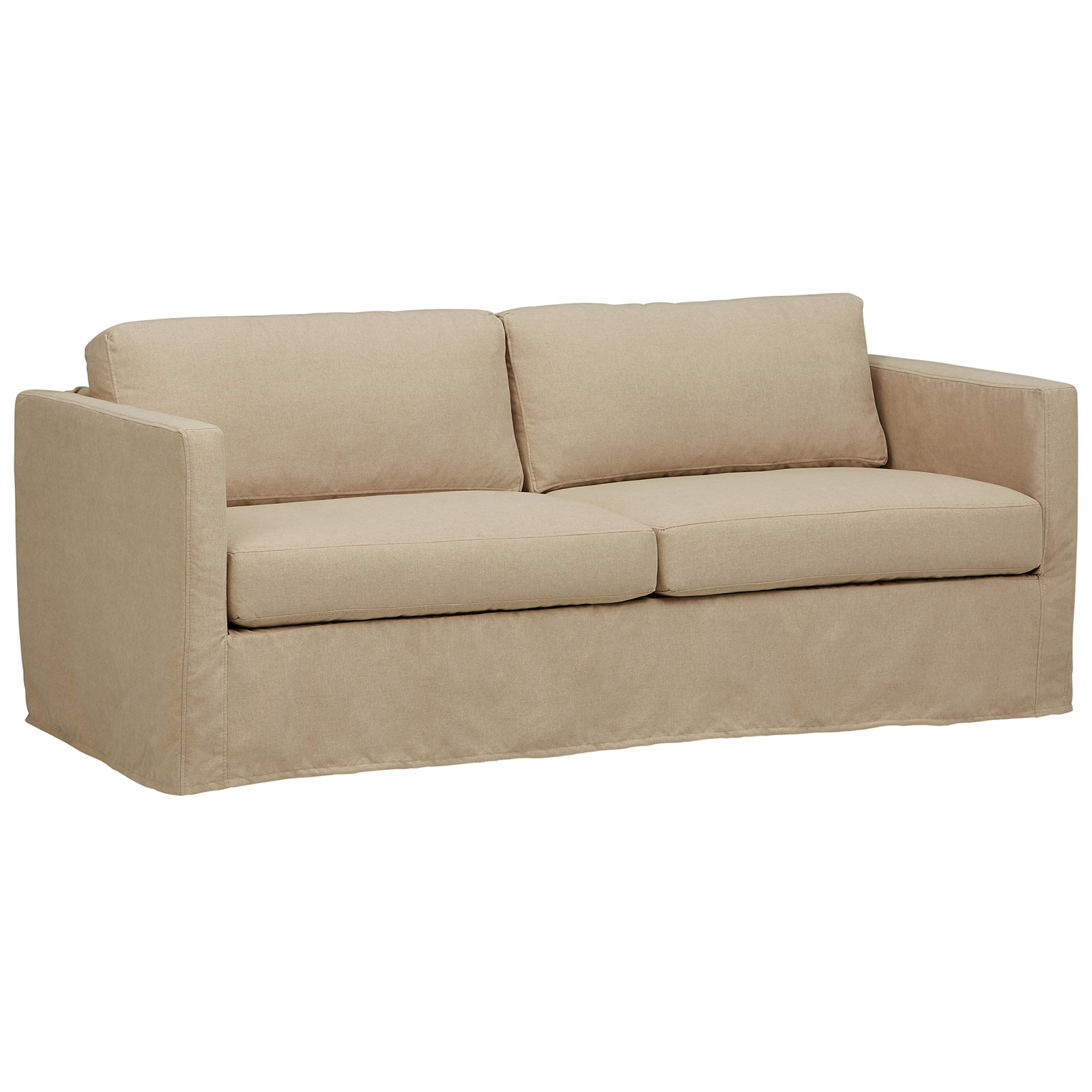}
    \caption{Target image.}
    \label{fig:white_bg_shot}
  \end{subfigure}
  \hfill
  \begin{subfigure}[b]{0.49\linewidth}
    \centering
    \includegraphics[width=3.6cm]{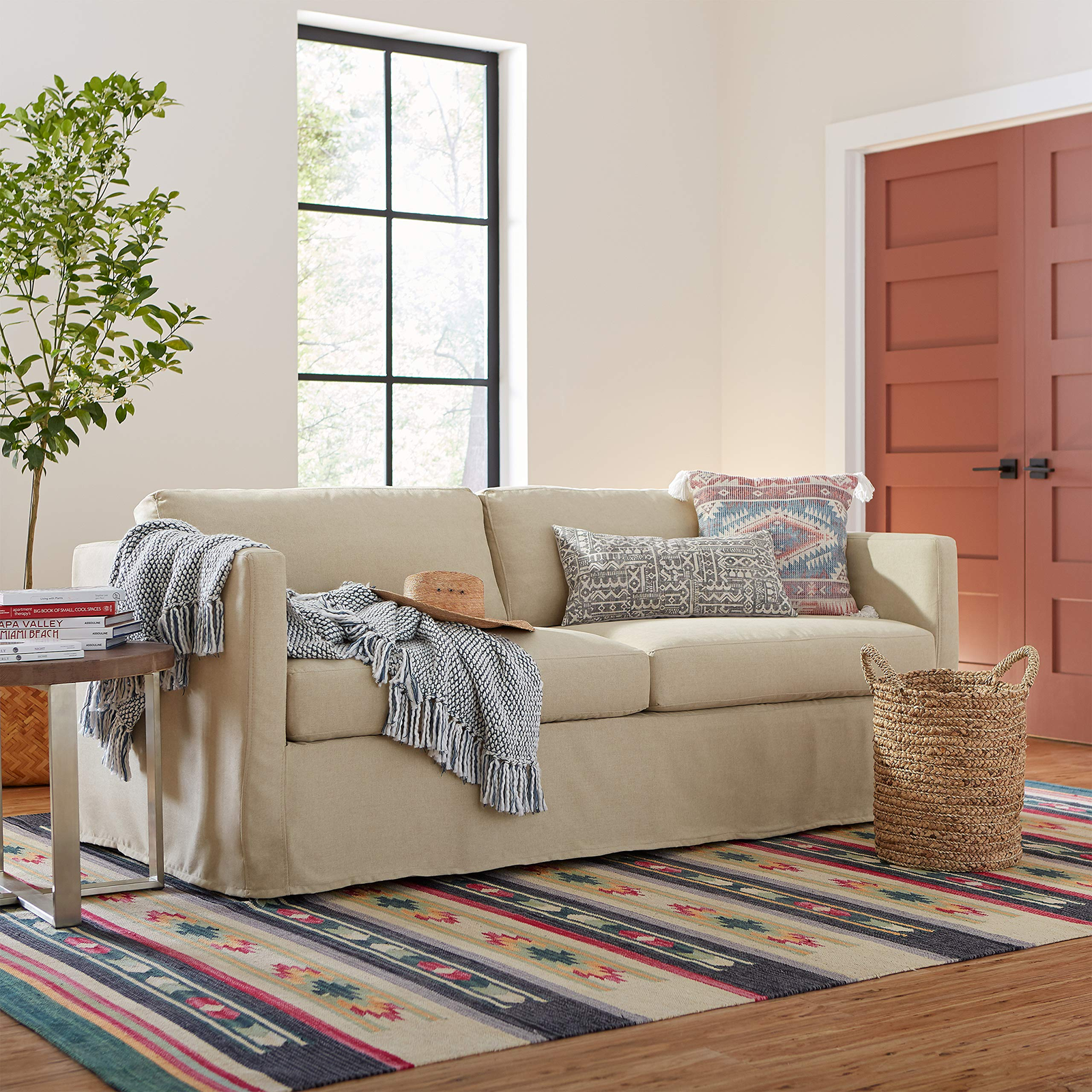}
    \caption{Milieu image.}
    \label{fig:milieu_shot}
  \end{subfigure}
  \caption{Example product image pair: white-background target (a) and milieu context (b).}
  \label{fig:product_and_catalogue_shot}
\end{figure}

\paragraph{Resulting Evaluation Set.}
The final evaluation set contains 4{,}725 products, i.e.\ 3.2\% of the 147{,}702 listings in ABO; the large majority is excluded by the first rule alone, because they lack structured dimension fields, a white-background target image, or a paired milieu image. Each product has a target image on a neutral background and a milieu image showing it in an ambient setting (Figure~\ref{fig:product_and_catalogue_shot}), and verified dimensions in centimeters. As product category we keep ABO's \texttt{product\_type} field: the items span 113 types with a highly unbalanced distribution, in which \texttt{CHAIR} (815 items, 17.2\%) and \texttt{SOFA} (558, 11.8\%) alone make up 29\% of the set, followed by \texttt{HOME\_FURNITURE\_AND\_DECOR}, \texttt{TABLE}, and \texttt{LAMP}, while 18 types occur only once. The types differ widely in size and homogeneity (e.g.\ sofas 192.2 $\pm$ 37.6 cm longest side, lamps 76.9 $\pm$ 50.0 cm). A Python script and index to reconstruct this exact evaluation set from ABO are available on Zenodo \citep{serban2026abo}.

\subsection{Models}
\label{subsec:models}
We evaluate MiniMax~M3, Gemini~3.8~Flash, Grok-4.6, and GPT-5.6~Terra through the OpenRouter API, which lets one harness sweep multiple providers under a shared request format and reasoning-control mechanism. Kimi K3, Claude Fable 5 and GLM-5V-Turbo were excluded from the full sweep for cost reasons: the four active models already consume about 196 million tokens (Table~\ref{tab:token-usage}), dominated by the image-heavy scene-context condition, and we prioritized covering both conditions on every model over covering more models in one condition. All requests use a temperature of 0.0.

\subsection{Prompt Construction}
\label{sec:prompt-construction}
\paragraph{Reasoning Control.}
For every request we disable the model's native reasoning (“thinking“) mode via OpenRouter's unified \texttt{reasoning} field. Two endpoints cannot be fully disabled this way: Gemini~3.8~Flash rejects the field with a “mandatory-reasoning error“ and was set to the provider's \textit{minimal} effort tier, and Grok-4.6 was set to its \textit{low} tier. MiniMax~M3 and GPT-5.6~Terra accept the full disable, which Section~\ref{subsec:token-usage} verifies from the usage data.

\paragraph{Prompt Structure.}
Since native reasoning is disabled, we keep the prompt deliberately simple and fixed and vary only one factor: whether the model receives the milieu image in addition to the product cutout (Figure~\ref{fig:prompt}). The model is told it is an image annotator estimating real-world object sizes and must return exactly three values in centimeters (width, length, height) as JSON; no reference-object guidance or intermediate reasoning steps are included, reflecting the finding in Section~\ref{sec:prelim-exploration} that such components produced no measurable benefit. In the scene-context condition, a target-object rule instructs the model to estimate only the object matching the cutout. We cannot verify from the three-number output that this rule was always followed, but the manual verification ensures that the milieu image contains the target product, and the per-category results (Section~\ref{subsec:per-category}) show no systematic degradation for types whose milieu images contain many other objects, such as lamps. The same prompt is used for all models without model-specific tuning, since per-model wording would confound model and prompt differences, and even minor presentation details can reorder VLM leaderboards \citep{feng2026visuallypromptedbenchmarkssurprisingly}. All $4 \times 2 \times 4{,}725 = 37{,}800$ requests returned a parseable three-value JSON answer.

\begin{figure}
\centering
    \begin{prompttemplate}{blue!25}
    \raggedright
    \promptheader{ROLE}
    You are an image annotator estimating
    real-world object sizes.\\[1mm]
    \promptheader{INPUT}
    1. A target image (cutout) of a
    specific object.\\[1mm]
    \textcolor{blue}{2. A milieu image (scene) containing
    instance(s) of that object.}\\[1mm]
    \promptheader{OBJECTIVE}
    Estimate the real-world size (in
    centimeters) of the object shown in the target image \\[1mm]
    \promptheader{RULES}
    \textcolor{blue}{1. Target Object: Estimate size only for objects that
    match the target object. Do not estimate unrelated objects.} \\[1mm]
    2. Dimensions: Report exactly three size values
    (in centimeters) for the object's main dimensions,
    in the order width, length, height. \\[1mm]
    \promptheader{OUTPUT FORMAT}
    Return ONLY valid JSON:

    '''json
    \{
      "sizes": [
        <number in cm>
      ]
    \}
    '''
    \end{prompttemplate}
\caption{Prompt used in both conditions. The blue lines are present only in the scene-context condition; in the no-context condition the rule numbering starts at 1.}
\label{fig:prompt}
\end{figure}

\subsection{Metrics}
\label{subsec:Metrics}
For each item we compute the absolute and signed error between estimated and ground-truth longest side, and identically for the volume. We report mean absolute error (MAE), median absolute error, mean absolute relative error (MARE, per-item absolute error divided by ground truth before averaging), its signed counterpart MRE, which reveals systematic over- or underestimation, Pearson's $r$ and Spearman's $\rho$ between estimated and true sizes, and the percentage of estimates within $\pm$5\% and $\pm$10\% of ground truth.

\subsection{Baselines}
\label{subsec:baselines}
To quantify how much accuracy category knowledge alone yields, we add two image-free baselines. The \emph{category-median} baseline predicts for each item the median ground-truth longest side (and volume) of all \emph{other} items with the same ABO \texttt{product\_type} (leave-one-out). The type is ABO's \texttt{product\_type} field (113 types among our items, Section~\ref{sec:dataset-adjustments}), which is part of the replication kit; the 18 singleton types fall back to the global median. The baseline is strong in that it is computed on the evaluation set itself, but it has no access to the image. The \emph{global-median} baseline predicts the median of all items. As a sensitivity check, the category baseline on ABO's finest browse-node leaves combined with \texttt{product\_type} (921 classes, singletons falling back to the type median) improves its MAE only from 25.9 to 21.0 cm, so the conclusions do not depend on the taxonomy's granularity.

\section{\uppercase{Results}}
\label{sec:results}
Table~\ref{tab:results} reports our headline accuracy metrics for
all four models under both context conditions. Gemini~3.8~Flash is
the strongest model overall, achieving the lowest MAE in both the
no-scene (10.02 cm) and scene (8.23 cm) conditions, and the highest
Pearson correlation with ground truth (up to $r=0.964$).
GPT-5.6~Terra is the clear outlier at the opposite end, with an MAE
roughly 1.5--1.9$\times$ higher than any other model (19.24 cm
no-scene, 15.82 cm scene) and the weakest correlation with ground
truth ($r=0.896$--$0.928$). Adding the milieu image as scene context
reduces MAE and MARE for every model, with the largest relative gains for Gemini~3.8~Flash and GPT-5.6~Terra (MAE reductions of approximately 18\%) and smaller gains for
Grok-4.6 and MiniMax~M3 (8--10\%). These reductions exceed the sampling uncertainty: the bootstrap 95\% confidence intervals of the MAE (1{,}000 resamples over items) are disjoint between the two conditions for every model (e.g.\ 9.58--10.46 cm vs.\ 7.86--8.65 cm for Gemini~3.8~Flash, 18.67--19.86 cm vs.\ 15.35--16.33 cm for GPT-5.6~Terra).
\begin{table*}[t]
\centering
\scriptsize
\setlength{\tabcolsep}{2pt}
\caption{Size-estimation longest side accuracy with volume equivalents (defined in Section~\ref{sec:dataset-adjustments}) in parentheses ($N=4{,}725$). MARE and signed MRE in \%; Mean/Median AE in cm (volume: m\textsuperscript{3}). Gemini 3.8F = Flash; GPT-5.6T = Terra. The last row is the image-free leave-one-out category-median baseline of Section~\ref{subsec:baselines}.}
\label{tab:results}
\begin{tabular*}{\textwidth}{@{\extracolsep{\fill}}llrrrrrrrr@{}}
\toprule
\textbf{Model} & \multicolumn{1}{c}{\textbf{Context}} & \multicolumn{1}{c}{\textbf{MARE}} & \multicolumn{1}{c}{\textbf{Mean AE}} & \multicolumn{1}{c}{\textbf{Median AE}} & \multicolumn{1}{c}{\textbf{MRE}} & \multicolumn{1}{c}{\textbf{Pearson $r$}} & \multicolumn{1}{c}{\textbf{Spearman $\rho$}} & \multicolumn{2}{c}{\textbf{Within (\%)}} \\
\cmidrule(lr){9-10}
& & & & & & & & \multicolumn{1}{c}{\textbf{$\le$5\%}} & \multicolumn{1}{c}{\textbf{$\le$10\%}} \\ \midrule
\textbf{Gemini 3.8F} & No Scene & 11.83 (57.93) & 10.02 (0.17) & 5.10 (0.04) & +4.62 (+40.75)  & 0.954 (0.838) & 0.960 (0.966) & 45.42 (18.58) & 67.72 (33.21) \\
\textbf{Gemini 3.8F} & Scene    &  9.14 (45.03) &  8.23 (0.14) & 4.50 (0.03) & +2.43 (+29.49)  & 0.964 (0.889) & 0.977 (0.981) & 50.84 (21.93) & 74.22 (39.20) \\ \addlinespace
\textbf{Grok-4.6}    & No Scene & 13.16 (61.95) & 11.33 (0.18) & 6.98 (0.04) & +6.19 (+43.45)  & 0.952 (0.761) & 0.962 (0.969) & 37.06 (15.09) & 59.77 (28.55) \\
\textbf{Grok-4.6}    & Scene    & 11.92 (57.91) & 10.43 (0.16) & 6.25 (0.04) & +5.61 (+40.80)  & 0.960 (0.878) & 0.969 (0.974) & 38.86 (16.21) & 63.24 (30.62) \\ \addlinespace
\textbf{MiniMax M3}  & No Scene & 13.72 (72.87) & 12.05 (0.18) & 6.81 (0.05) & +5.28 (+50.66)  & 0.950 (0.777) & 0.951 (0.957) & 37.54 (14.05) & 59.22 (26.73) \\
\textbf{MiniMax M3}  & Scene    & 12.65 (60.29) & 10.87 (0.18) & 6.66 (0.05) & +4.57 (+39.63)  & 0.959 (0.766) & 0.958 (0.967) & 36.80 (13.10) & 61.04 (26.33) \\ \addlinespace
\textbf{GPT-5.6T}   & No Scene & 25.42 (155.04) & 19.24 (0.23) & 12.29 (0.08) & +18.93 (+144.65) & 0.896 (0.811) & 0.895 (0.929) & 25.29 (9.88)  & 41.42 (20.08) \\
\textbf{GPT-5.6T}   & Scene    & 20.79 (111.99) & 15.82 (0.19) & 10.56 (0.06) & +15.08 (+100.77) & 0.928 (0.816) & 0.932 (0.957) & 26.84 (11.85) & 45.19 (22.52) \\ \midrule
\textbf{Category median} & -- & 30.36 (279.29) & 25.90 (0.28) & 16.19 (0.12) & +9.52 (+240.54) & 0.775 (0.785) & 0.740 (0.808) & 19.32 (9.42) & 33.46 (14.14) \\ \bottomrule
\end{tabular*}
\end{table*}

Two further patterns in Table~\ref{tab:results} matter for the interpretation. First, median AE is consistently far below MAE (e.g.\ 4.50 vs.\ 8.23 cm for Gemini~3.8~Flash with scene): the error distribution is heavy-tailed, a typical item is within a few centimeters of the true size, and a minority of large misses dominates the mean. Second, MRE is positive for every model-condition pair, i.e.\ all models tend to overestimate; the bias shrinks with scene context, most sharply for GPT-5.6~Terra (+18.9\% to +15.1\%), but never vanishes. The volume correlations are substantially lower than their longest-side counterparts (Pearson 0.761--0.889 vs.\ 0.896--0.964), showing that compounding the per-axis errors exposes a much weaker relationship than the single-axis reduction suggests. Finally, even the best configuration (Gemini~3.8~Flash with scene, 74\% within $\pm$10\%) misses this practically relevant tolerance band on roughly one in four items.

\subsection{Comparison with the Category Baselines}
\label{subsec:baseline-results}
The last row of Table~\ref{tab:results} shows that category knowledge alone is far from sufficient. The leave-one-out category median reaches an MAE of 25.90 cm (MARE 30.4\%, $\rho = 0.740$, 33.5\% within $\pm$10\%), the global median 47.58 cm (MARE 70.9\%). Every model in every condition is substantially better: even the weakest configuration, GPT-5.6~Terra without scene, has a 26\% lower MAE than the category baseline, and the best, Gemini~3.8~Flash with scene, a 68\% lower MAE with more than twice the within-10\% rate. On the item level, Gemini~3.8~Flash beats the category median on 72\% (no scene) and 76\% (scene) of all items, Grok-4.6 and MiniMax~M3 on 68--69\%, and GPT-5.6~Terra on 55--59\%. The VLMs therefore extract substantial item-specific size information from the image.

\subsection{Per-Category Results}
\label{subsec:per-category}
Table~\ref{tab:per-category} breaks the longest-side MAE down by the ten most frequent ABO product types (73\% of the evaluation set). Three patterns stand out. First, the margin over the baseline is largest for heterogeneous types whose members vary widely in size (\texttt{HOME\_FURNITURE\_AND\_DECOR}, \texttt{TABLE}, \texttt{LAMP}, \texttt{HOME}): baseline errors of 31--40 cm shrink to 7--15 cm for the three stronger models, so the models clearly size the individual object rather than the type. For large canonical objects with a tight prior (\texttt{SOFA}, \texttt{BED}) the margin is smaller, and GPT-5.6~Terra barely beats the baseline on sofas (28.1 and 26.1 vs.\ 28.5 cm). Second, the benefit of scene context is strongly type-dependent: consistent across models for tables and the heterogeneous home-furniture types, where the milieu image resolves whether an object is a side table or a dining table, dramatic in individual cases such as GPT-5.6~Terra on \texttt{STORAGE\_HOOK} (42.4 to 14.0 cm, wall hook vs.\ free-standing coat rack), absent for beds, and mixed for sofas, lamps, and chairs; MiniMax~M3 even degrades with context on lamps, ottomans, and storage hooks. Third, the effect is not uniform across items: with scene context, Gemini~3.8~Flash improves on 46\% of the items and worsens on 35\% (Grok-4.6: 35/30\%, MiniMax~M3: 32/30\%, GPT-5.6~Terra: 43/35\%), while the remaining 19--38\% of the items receive an identical estimate in both conditions. The scene image thus helps on balance by removing large errors, not by refining every estimate, and a sizable share of estimates is not influenced by the additional image at all, consistent with prior-driven estimation for those items (Section~\ref{sec:discussion}).

\begin{table*}[t]
\centering
\scriptsize
\setlength{\tabcolsep}{3pt}
\caption{Longest-side MAE in cm per ABO \texttt{product\_type} for the ten most frequent types (\emph{n} = items, GT = mean ground-truth longest side in cm), for the leave-one-out category-median baseline and for each model without / with scene context. Bold marks the better of the two conditions per model where the difference exceeds 0.5 cm.}
\label{tab:per-category}
\begin{tabular*}{\textwidth}{@{\extracolsep{\fill}}lrrrrrrr@{}}
\toprule
\textbf{Product type} & \textbf{\emph{n}} & \textbf{GT} & \textbf{Baseline} & \textbf{Gemini 3.8F} & \textbf{Grok-4.6} & \textbf{MiniMax M3} & \textbf{GPT-5.6T} \\
 & & & & no / scene & no / scene & no / scene & no / scene \\ \midrule
\texttt{CHAIR}                    & 815 &  95.1 & 11.86 &  8.13 / \textbf{4.59} &  6.25 / 6.39 &  6.01 / 6.03 &  9.37 / 9.25 \\
\texttt{SOFA}                     & 558 & 192.2 & 28.50 & 18.43 / 18.49 & 20.87 / \textbf{20.26} & 20.17 / \textbf{16.84} & 28.07 / \textbf{26.07} \\
\texttt{HOME\_FURNITURE\_AND\_DECOR} & 502 & 133.3 & 40.35 & 10.90 / \textbf{9.27} & 13.40 / \textbf{12.35} & 14.36 / \textbf{13.00} & 20.91 / \textbf{17.16} \\
\texttt{TABLE}                    & 441 &  95.4 & 36.22 &  9.18 / \textbf{7.11} & 10.12 / \textbf{9.04} & 11.68 / \textbf{9.53} & 22.61 / \textbf{14.89} \\
\texttt{LAMP}                     & 364 &  76.9 & 33.97 &  9.63 / \textbf{8.88} & 12.14 / 12.30 & \textbf{10.71} / 13.33 & 21.03 / \textbf{19.23} \\
\texttt{STOOL\_SEATING}           & 185 &  93.8 & 13.23 &  7.33 / \textbf{5.03} &  5.92 / 5.68 &  6.98 / 6.91 & 10.02 / \textbf{8.13} \\
\texttt{OTTOMAN}                  & 168 &  89.4 & 28.25 &  7.42 / \textbf{6.16} &  9.17 / \textbf{7.78} & \textbf{9.71} / 12.11 & 20.93 / \textbf{15.75} \\
\texttt{HOME}                     & 154 &  55.3 & 30.94 &  9.41 / \textbf{7.46} &  9.50 / \textbf{8.51} & 11.67 / \textbf{10.34} & 22.72 / \textbf{19.08} \\
\texttt{BED}                      & 130 & 205.7 & 18.24 &  7.88 / \textbf{7.09} & 10.86 / 11.07 &  9.33 / 9.42 & \textbf{8.02} / 10.05 \\
\texttt{STORAGE\_HOOK}            & 109 &  62.5 & 20.88 & \textbf{4.55} / 5.40 & \textbf{11.96} / 13.06 & \textbf{10.45} / 15.72 & 42.44 / \textbf{14.02} \\ \bottomrule
\end{tabular*}
\end{table*}

\subsection{Token Usage}
\label{subsec:token-usage}
Table~\ref{tab:token-usage} reports token consumption across the full sweep. Adding the milieu image roughly doubles input tokens for three models ($1.9$--$2.1\times$), consistent with one additional image per request; MiniMax~M3 is an outlier at $10.9\times$ (845.7 to 9{,}193.3 tokens per request), which we could not attribute to image content and flag as a likely provider-side image-tokenization artifact of this endpoint. Despite the disable request, Gemini~3.8~Flash and Grok-4.6 report that reasoning accounts for 86.7--91.7\% and 96.1--97.0\% of their output tokens at their respective effort floors, whereas MiniMax~M3 and GPT-5.6~Terra report zero reasoning tokens, confirming that reasoning was fully suppressed for the latter two.

\begin{table*}[t]
\centering
\scriptsize
\caption{Total token usage across the full 4{,}725-item sweep, by model and condition. Reasoning tokens are billed despite our disable request and are included within the total output-token count, not additional to it. Gemini~3.8~Flash and Grok-4.6 could not be fully disabled and ran at their lowest available reasoning-effort tier instead (Section~\ref{sec:prompt-construction}).}
\begin{tabular*}{\textwidth}{@{\extracolsep{\fill}}llrrrr@{}}
\toprule
\textbf{Model} & \textbf{Context} & \textbf{Input Tokens} & \multicolumn{2}{c}{\textbf{Output Tokens}} & \textbf{Total Tokens} \\
\cmidrule(lr){4-5}
& & & \textbf{Combined} & \textbf{Reasoning} & \\ \midrule
\textbf{Gemini 3.8 Flash} & No Scene & 6{,}138{,}728  & 1{,}719{,}309 & 1{,}575{,}928 & 7{,}858{,}037  \\
\textbf{Gemini 3.8 Flash} & Scene    & 11{,}783{,}221 & 1{,}455{,}727 & 1{,}262{,}415 & 13{,}238{,}948 \\ \addlinespace
\textbf{Grok-4.6}         & No Scene & 12{,}632{,}915 & 1{,}897{,}535 & 1{,}841{,}345 & 14{,}530{,}450 \\
\textbf{Grok-4.6}         & Scene    & 24{,}053{,}511 & 2{,}547{,}197 & 2{,}448{,}982 & 26{,}600{,}708 \\ \addlinespace
\textbf{MiniMax M3}       & No Scene & 3{,}995{,}726  & 101{,}951   & 0             & 4{,}097{,}677  \\
\textbf{MiniMax M3}       & Scene    & 43{,}438{,}519 & 140{,}890   & 0             & 43{,}579{,}409 \\ \addlinespace
\textbf{GPT-5.6 Terra}    & No Scene & 27{,}855{,}523 & 99{,}615    & 0             & 27{,}955{,}138 \\
\textbf{GPT-5.6 Terra}    & Scene    & 58{,}283{,}718 & 129{,}323   & 0             & 58{,}413{,}041 \\ \bottomrule
\end{tabular*}
\label{tab:token-usage}
\end{table*}
\section{\uppercase{Discussion and Limitations}}
\label{sec:discussion}
\subsection{Preliminary Prompt-Component Exploration}
\label{sec:prelim-exploration}
Before settling on the minimal two-condition design, we explored a one-factor-at-a-time sweep over two additional prompt components, building on reference-object prompting \citep{liao2024qspatial,alker2025hail}. The first added a guidance block listing common household objects with their standard dimensions (e.g.\ a dining table at $\sim$76\,cm, a door at $\sim$200\,cm height). The second replaced direct size reporting with an explicit, checkable computation: the model identified a reference object in the milieu image, stated its standard size $S_R$, and estimated the visual size ratio $r$ between target and reference; the estimate was then computed programmatically as $S_R \cdot r$. Across repeated runs, neither component, individually or combined, changed estimation accuracy by more than the run-to-run variation of the identical condition. We had expected an externalized, checkable structure to at least partially substitute for the disabled internal reasoning; repeating the exploration with native reasoning enabled gave the same null result, so it is not an artifact of reasoning being disabled. We report this as a negative result: it runs counter to prior findings on reference-object prompting obtained with reasoning enabled \citep{liao2024qspatial,alker2025hail}, but is in line with VSI-Bench, where chain-of-thought and self-consistency likewise failed to improve spatial answers \citep{11094671}.

\subsection{Category Priors vs.\ Genuine Visual Measurement}
Are the models measuring the object, or mostly relying on category-level size priors learned during pretraining (“sofas are about 2\,m wide“) and adjusting only slightly? Three patterns in Section~\ref{sec:results} point toward heavy reliance on priors: the heavy-tailed errors behind the mean/median gap; the much lower volume correlations, consistent with anchoring on a plausible single dimension rather than recovering all three axes; and the consistently positive MRE that shrinks with scene context but never reaches zero, as if the milieu image only partly corrects an underlying prior. The category baselines bound this interpretation from the other side: a predictor that knows only the product type has a 1.3--3.1$\times$ higher MAE than the VLMs, is worse on every metric, and is beaten on 55--76\% of the individual items; refining that prior to 921 fine-grained classes (Section~\ref{subsec:baselines}) still leaves it at 21 cm, far from the 8--19 cm of the models. The estimates are therefore not reducible to a category prior; a more accurate description is a strong prior adjusted by item-specific visual evidence, with the adjustment being too small (positive MRE) and occasionally absent (identical estimates with and without scene for 19--38\% of the items). This matches \citet{meivar2026illposeddesignprobingevidence}, who find target identity to be the most load-bearing cue for VLM size estimates and global scene geometry to be largely unused.

\subsection{Why Do the Models Differ?}
\label{sec:model-differences}
Our data cannot attribute model differences to architecture or training data, which are undisclosed, but three observations narrow them down. First, the ranking is stable across conditions and metrics: Gemini~3.8~Flash leads and GPT-5.6~Terra trails everywhere, so the model effect is larger than, and independent of, the context effect. Second, the deficit of GPT-5.6~Terra is not a calibration offset: subtracting its mean signed error lowers its MAE only from 19.24 to 18.18 cm, and dividing its estimates by its median estimate-to-truth ratio of 1.08 only to 18.32 cm, still far from the 10--12 cm of the other models. Its errors are heavy-tailed and item-specific, which points to a weaker or less reliably retrieved size prior rather than a different unit convention. Third, the two strongest models, Gemini~3.8~Flash and Grok-4.6, are exactly the two whose reasoning could not be switched off and which still spend 87--97\% of their output tokens on internal reasoning (Table~\ref{tab:token-usage}). This confound is inherent to current provider APIs, but it is consistent with the interpretation that some internal deliberation helps size estimation while externalized reasoning structure in the prompt (Section~\ref{sec:prelim-exploration}) does not. Why scene context helps is easier to explain: the milieu image restores the cues that product photography removes, a ground plane, walls, doors, and neighboring furniture of known typical size, i.e.\ the reference objects that prior work exploits through explicit prompting \citep{liao2024qspatial,alker2025hail}. The models use these cues even without being asked to, and the gain is largest for the two models with the largest no-scene error. It is bounded, however, because a single uncalibrated milieu image constrains relative, not absolute, scale, and because the benefit varies strongly between product types (Table~\ref{tab:per-category}).

\subsection{Limitations}
\label{sec:single-image}
ABO spans 576 product types~\citep{collins2022abo}, but our filtering did not affect all categories equally: discarding adjustable, variant-dependent, and bundled listings removes more furniture and multi-piece items than single-configuration products, so the evaluation set does not preserve ABO's category mix, and its 113 product types are highly unbalanced, with chairs and sofas alone making up 29\% of the items and dominating the aggregate metrics in Table~\ref{tab:results}. ABO is also a single retailer's catalog, so our findings may not transfer to other photographic conventions or to categories such as groceries and apparel. Finally, recovering absolute scale from a single 2D image is fundamentally ill-posed: the same silhouette is consistent with infinitely many sizes at different distances and focal lengths, and product photography strips away the scene cues that could resolve it. The milieu image supplies such cues, and the consistent improvement it produces confirms that this is a real bottleneck, but it carries no calibration information of its own, so the heavy-tailed outliers behind the mean/median gap likely correspond to items where neither condition gave the model a reliable scale cue.

\section{\uppercase{Conclusion}}
\label{sec:conclusion}
We benchmarked four commercial VLMs on metric size estimation from a curated, manually verified 4{,}725-item subset of ABO, comparing a product-cutout-only condition against one adding scene context, with native reasoning disabled where the API permits it. Four findings emerge. (i) Off-the-shelf VLMs estimate the longest side of household products with an MAE of 8--19 cm and a median error of 4.5--12 cm, and they do so from the image, not from the product type alone: a leave-one-out category-median baseline has a 1.3--3.1$\times$ higher MAE and is beaten on 55--76\% of the items. (ii) Scene context is a reliable lever: the milieu image reduces MAE and MARE for every model, with non-overlapping bootstrap confidence intervals, whereas elaborate prompt components yielded no benefit distinguishable from run-to-run noise, with and without native reasoning. (iii) The benefit of context is type-specific, large for objects whose scale is ambiguous from the cutout and absent for objects with a tight size prior, and 19--38\% of the estimates do not change at all when the scene image is added. (iv) The models differ more than the conditions do: Gemini~3.8~Flash leads on every metric and GPT-5.6~Terra trails throughout, with item-specific errors that neither an additive nor a multiplicative calibration removes. Together, the results characterize current VLM size estimation as a strong category prior that is adjusted, but not replaced, by visual evidence. The accuracy of the best configuration (74\% of items within $\pm$10\%) points towards a concrete application: an automatic size-consistency check for composited or generated milieu images. Future work should add geometry-based estimators as baselines, extend the comparison to open-weight models whose reasoning can be controlled fully, and test whether multiple milieu views narrow the remaining gap.

\section*{\uppercase{Acknowledgements}}
This work was carried out in the project \emph{AI Fotostudio}, funded by the Austrian Research Promotion Agency (FFG, grant no.\ 60931359) in cooperation with Premedia GmbH. The evaluation set was created during this study and can be reconstructed with the replication kit on Zenodo \citep{serban2026abo}. The authors have no conflicts of interest to declare. AI tools were used for text manipulations such as proofreading, rewriting, and shortening the text.

\bibliographystyle{apalike}
{\small
\bibliography{references}}

\end{document}